\documentclass[conference]{IEEEtran}
\IEEEoverridecommandlockouts
\usepackage{cite}
\usepackage{amsmath,amssymb,amsfonts}
\usepackage{algorithmic}
\usepackage{graphicx}
\usepackage{textcomp}
\usepackage{xcolor}
\usepackage{bm}
\usepackage{verbatim}
\usepackage{booktabs}
\usepackage{multirow}
\usepackage{hyperref}
\usepackage{scrextend}
\usepackage{enumerate}
\def\BibTeX{{\rm B\kern-.05em{\sc i\kern-.025em b}\kern-.08em
    T\kern-.1667em\lower.7ex\hbox{E}\kern-.125emX}}
\begin{document}

\title{Semi-decentralized Training of Spatio-Temporal Graph Neural Networks for Traffic Prediction\\
\thanks{This work has been supported by the Horizon Europe WIDERA program under the grant agreement No. 101079214 (AIoTwin). We acknowledge the EuroHPC Joint Undertaking for awarding this project access to the EuroHPC supercomputer LEONARDO, hosted by CINECA (Italy) and the LEONARDO consortium through an EuroHPC Development Access call.}
}

\author{
    Ivan Kralj\IEEEauthorrefmark{1}, 
    Lodovico Giaretta\IEEEauthorrefmark{2}, 
    Gordan Ježić\IEEEauthorrefmark{1},
    Ivana Podnar Žarko\IEEEauthorrefmark{1},
    Šarūnas Girdzijauskas\IEEEauthorrefmark{2}\IEEEauthorrefmark{3}\\
    \IEEEauthorrefmark{1}University of Zagreb, Faculty of Electrical Engineering and Computing, Croatia \\
    \{ivan.kralj, gordan.jezic\}@fer.hr\\
    \IEEEauthorrefmark{2}RISE Research Institutes of Sweden, Sweden \\
    \IEEEauthorrefmark{3}KTH Royal Institute of Technology, Sweden
}

\begin{comment}
    \author{\IEEEauthorblockN{Ivan Kralj}
\IEEEauthorblockA{\textit{Department of Telecommunications} \\
\textit{University of Zagreb,}\\
\textit{Faculty of Electrical}\\
\textit{Engineering and Computing}\\
Zagreb, Croatia \\
ivan.kralj@fer.hr}
\and
\IEEEauthorblockN{Lodovico Giaretta}
\IEEEauthorblockA{\textit{Department of Computer Science} \\
\textit{RISE Research Institutes of Sweden}\\
Kista, Sweden \\
lodovico.giaretta@ri.se}
\and
\IEEEauthorblockN{Gordan Ježić}
\IEEEauthorblockA{\textit{Department of Telecommunications} \\
\textit{University of Zagreb,}\\
\textit{Faculty of Electrical}\\
\textit{Engineering and Computing}\\
Zagreb, Croatia \\
gordan.jezic@fer.hr}
}
\end{comment}

\maketitle

\begin{abstract}
In smart mobility, large networks of geographically distributed sensors produce vast amounts of high-frequency spatio-temporal data that must be processed in real time to avoid major disruptions. Traditional centralized approaches are increasingly unsuitable to this task, as they struggle to scale with expanding sensor networks, and reliability issues in central components can easily affect the whole deployment. To address these challenges, we explore and adapt semi-decentralized training techniques for Spatio-Temporal Graph Neural Networks (ST-GNNs) in the smart mobility domain. We implement a simulation framework where sensors are grouped by proximity into multiple cloudlets, each handling a subgraph of the traffic graph, fetching node features from other cloudlets to train its own local ST-GNN model, and exchanging model updates with other cloudlets to ensure consistency, enhancing scalability and removing reliance on a centralized aggregator. We perform extensive comparative evaluation of four different ST-GNN training setups---centralized, traditional FL, server-free FL, and Gossip Learning---on large-scale traffic datasets, the METR-LA and PeMS-BAY datasets, for short-, mid-, and long-term vehicle speed predictions. Experimental results show that semi-decentralized setups are comparable to centralized approaches in performance metrics, while offering advantages in terms of scalability and fault tolerance. In addition, we highlight often overlooked issues in existing literature for distributed ST-GNNs, such as the variation in model performance across different geographical areas due to region-specific traffic patterns, and the significant communication overhead and computational costs. However, due to the planar nature of graphs, per-cloudlet costs remain consistent as the network grows, unlike the growing costs in a centralized approach.
\end{abstract}


\section{Introduction}

\begin{figure*}[ht!]
  \centering
  \includegraphics[width=1\linewidth]{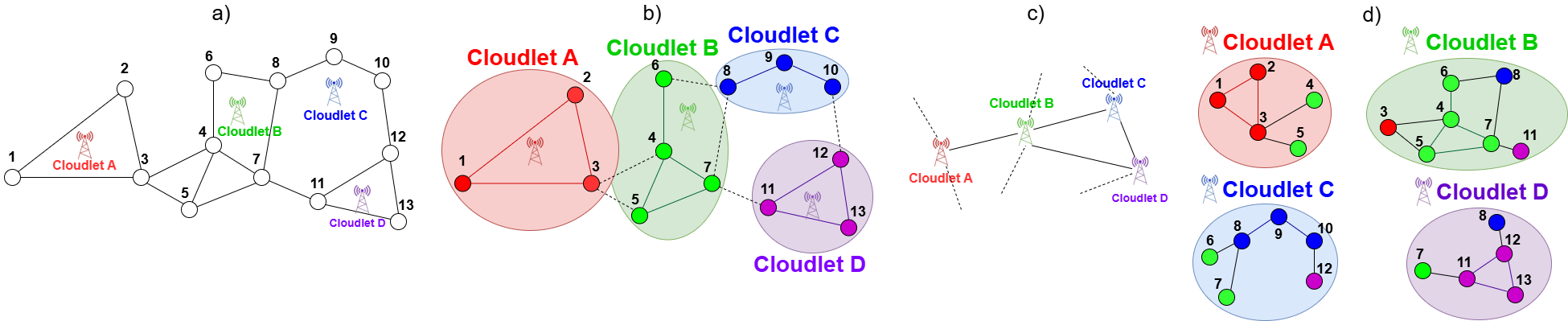}
  \caption{Graph partitioning and communication. a) Geographically distributed sensor network and base stations b) Graph partitioning of the sensors into cloudlets based on geographical proximity. c) Cloudlet-to-cloudlet communication network for exchanging node features and model updates. d) After communicating with neighbouring cloudlets, each cloudlet can construct the ST-GNN subgraph required for training on its local nodes}
  \label{fig:0.5}
\end{figure*}

Smart mobility is integral to urban planning, particularly in transportation systems, influencing development and traffic management \cite{cite_1}. Traffic prediction is a key component of smart mobility \cite{cite_2, cite_3, cite_4, cite_5, cite_6} and includes tasks such as vehicle speed and public transport forecasting. As demand for these services rises, the need for accurate and efficient traffic forecasting grows, enabling effective resource management, congestion reduction, and route optimization.

Like other prediction tasks, traffic forecasting methods can generally be divided into classical statistical methods like autoregressive integrated moving average (ARIMA) \cite{cite_34} and Machine Learning (ML) methods. The latter can be further split into ``traditional'' ML methods, such as gradient boosting \cite{cite_35}, and Deep Learning (DL)-based methods including Long Short-Term Memory (LSTM) \cite{cite_7} and Diffusion Convolutional Recurrent Neural Networks (DCRNN) \cite{cite_8}. However, all of these methods only exploit temporal dependencies, overlooking spatial ones.

Graph Neural Networks (GNNs) \cite{cite_29} address these issues by modeling traffic data as a graph, where nodes represent physical locations, and edges spatial connections between these locations. Specifically, Spatio-Temporal Graph Neural Networks (ST-GNNs) \cite{cite_16} extend GNNs to capture both spatial and temporal dependencies, making them more effective than other DL-based approaches for traffic forecasting.  Thus, numerous ST-GNN architectures have been developed and tested for this task \cite{cite_16}.

However, most previous works have focused on ST-GNN trained in centralized environments, where data from all sensors is continuously collected. But the task of collecting data, performing DL training and inference, and sending back commands is challenging to perform reliably in real-time in a centralized system. As sensor networks expand, scaling a central control system to handle the increased volume of data in real-time with low latency becomes increasingly difficult. Furthermore, any issues in the central control system can easily result in a complete interruption of the service, requiring significant investments to maximize reliability. Given the key role played by mobility infrastructure in our societies, a centralized system can also become a significant target for hostile actors and a major cyber-weakness in strategic infrastructure management.

In this paper, we explore and adapt distributed training techniques for ST-GNNs in the smart mobility domain through semi-decentralized methods at the network edge, ensuring scalability and reliability. This approach utilizes geographically distributed cloudlets, hosted on edge devices with moderate computation and communication capabilities, as shown in Figure~\ref{fig:0.5}-a. Each cloudlet manages a subgraph of the overall traffic graph (Fig. \ref{fig:0.5}-b), shares node features necessary for local training step and model updates with other cloudlets to maintain consistency (Fig. \ref{fig:0.5}-c), and constructs ST-GNN subgraph required for training on its local nodes (Fig. \ref{fig:0.5}-d). The contributions of this paper can be summarized as follows:

\begin{itemize}
\item To our knowledge, we are the first to investigate semi-decentralized ST-GNN training with geographically distributed cloudlets
\item We develop a simulation framework for semi-decentralized ST-GNN training across cloudlets, allowing us to test different approaches.
\item Using our framework, we implement and compare four different training approaches---centralized, traditional FL, server-free FL, and Gossip Learning---providing the first comprehensive analysis of the benefits and drawbacks of these approaches for our targeted use-case, assessing their suitability for real-world deployment and informing future research endeavours.
\item We evaluate these approaches on two real-world datasets, METR-LA and PeMS-BAY \cite{cite_31}, across short-, mid-, and long-term traffic predictions, with a detailed analysis of model performance, cloudlet-specific metrics, communication overhead, and computational costs.
\end{itemize}

Our experiments show that semi-decentralized approaches perform competitively against the centralized approach, especially for short-term prediction. Crucially, we provide insights on notable challenges in distributed ST-GNN training, which to our knowledge has not been explored in existing literature, such as highlighting the variation in model performance across different geographical areas due to region-specific traffic patterns, and the significant communication overhead and computational costs that arise from the large receptive field of GNNs, leading to substantial data transfers and increased computation of partial embeddings. However, communication overhead is mitigated by the planar structure of graphs, keeping per-cloudlet costs consistent as the network grows.
\section{Preliminaries}\label{sec:preliminares}
\subsection{Traffic prediction}
Traffic prediction is one of the most common and critical time-series forecasting tasks, as it estimates traffic metrics, such as speed, volume, and density to monitor current traffic conditions and predict future trends. It is generally classified into three forecasting horizons: short-term (15 min), mid-term (30 min), and long-term (60 min) \cite{cite_36}. In the context of ST-GNNs, the traffic system is represented as a spatial graph at the \textit{t}-th time step, denoted by $\mathcal{G}_t$ = ($\mathcal{V}_t$, $\mathcal{E}$, W), where $v_i, v_j\in \mathcal{V}_t$ is a finite set of nodes corresponding to observations from \textit{n} IoT-devices in the traffic network, $(v_i, v_j)\in \mathcal{E}$ is a set of edges connecting the IoT-devices, and $W\in \mathbb{R}^{n \times n}$ is a weighted adjacency matrix that encodes the spatial relationships between these IoT-devices. The spatial component is defined by the physical locations of IoT devices, while the temporal component arises from traffic features that evolve over time, influencing the graph's dynamics. While this work focuses on traffic prediction, methods for other smart mobility tasks can be generalized to this domain due to their shared reliance on spatio-temporal data. 

\subsection{Spatio-Temporal Graph Neural Networks}
ST-GNNs are DL models designed to capture both spatial and temporal dependencies in structured data, allowing it to simultaneously account for spatial correlations between nodes and temporal correlations over time, making it particularly suited for traffic prediction. By leveraging spatial dependencies, ST-GNNs provide a more accurate representation of how local traffic conditions at one location impact neighbouring locations.

Multiple ST-GNNs have been developed \cite{cite_16}, such as Graph Recurrent Network (GRN) \cite{cite_12}, Spatio-Temporal Graph Convolutional Network (ST-GCN) \cite{cite_13}, Graph Attention LSTM Network \cite{cite_14}, and many more \cite{cite_16}. Each technique addresses specific challenges or offers unique advantages over others. In this paper, we utilize ST-GCN due to their superior ability to handle traffic prediction tasks in the IoT context compared to other techniques. Experiments demonstrated that ST-GCN consistently outperforms other models in key evaluation metrics (MAE, MAPE, and RMSE) \cite{cite_13}.

\subsection{Traditional Federated Learning}
Traditional FL \cite{cite_30} is a collaborative machine learning (ML) approach where multiple clients work together to train a model, coordinated by a central aggregator. Instead of sharing raw data, which remains stored locally on each client device, the clients exchange only model updates with the server. These updates contain just the minimum information needed for the learning task, carefully scoped to limit the exposure of client data. The central aggregator then aggregates these focused updates as soon as they are received, ensuring that the learning process is both efficient and compliant with data minimization principles.

\subsection{Server-free Federated Learning}
Server-free FL \cite{cite_26} is a decentralized variant of the traditional FL approach. Unlike traditional FL, where a central aggregator is responsible for coordinating the model updates, server-free FL operates without a central entity. Instead, participating devices communicate directly with one another to exchange model parameters and perform local updates. After initializing their models, devices exchange model parameters with their neighbours, aggregate these received models with their own local model, and proceed to train on a subset of data. This iterative process allows for continuous refinement of models across the network, as the updated models are exchanged again between neighbouring devices.

\subsection{Gossip Learning}
Gossip Learning \cite{cite_27} is a decentralized protocol for training ML models in distributed settings without the need for a central coordinator. It is designed to be highly scalable and robust, allowing model updates to be propagated efficiently across a network of devices or nodes. Each device stores two models in its memory (as a FIFO buffer), typically representing the most recent models received from neighbouring devices. At each iteration of the gossip protocol, a device averages the weights of these two models to create an aggregated model. The device then performs one local training step using its own data, refining the aggregated model based on its local observations. Once the model is updated, the device forwards it to a randomly chosen device in the entire network. This process is repeated across the network, ensuring that models continuously evolve and improve as they traverse different nodes, collecting knowledge from the data they encounter at each location.

\subsection{Distributed training}
In distributed machine learning, training paradigms can be categorized based on their system architecture and coordination mechanisms:

\begin{itemize}
\item \textbf{Centralized Training}: Data is collected and processed on a single server or coordinator, which performs model updates.
\item \textbf{Semi-Decentralized Training}: A hybrid approach where devices are organized into clusters or hierarchies. Local coordination occurs within clusters, while limited global coordination ensures model consistency.
\item \textbf{Fully-Decentralized Training}: Devices collaborate in a peer-to-peer manner, exchanging model updates directly without relying on a central coordinator.
\end{itemize}
\section{Methodology}
\subsection{Scenario}
The traffic prediction scenario involves IoT-devices (i.e., sensors) deployed along highways at fixed locations to capture traffic metrics in real-time at fixed time intervals. To streamline data processing, IoT-devices are assigned to local base stations (BS) at fixed locations, referred to as cloudlets, based on their physical proximity and communication range. The communication between IoT-devices and its BS is facilitated using Low-Power Wide-Area Network (LPWAN) protocols, such as LoRa or NB-IoT, which enable efficient, long-range, and low-power data transmission. Base stations, in addition to serving as communication hubs, are equipped with moderate computational resources. This existing computational power can be leveraged for training ML models, such as ST-GNNs.

\subsection{Distributed Training Setups}
We adopt a semi-decentralized ST-GNN architecture inspired by Nazzal et al. \cite{cite_25}, as it's one of the most effective architectures for distributed training. Figure~\ref{fig:0.5} shows key components of such architecture. In Fig.~\ref{fig:0.5}-a, traffic network is represented as a graph, where nodes correspond to IoT-devices, and edges represent road connections based on physical proximity. In Fig.~\ref{fig:0.5}-b, the graph is partitioned into subgraphs based on geographical proximity, each managed by a local cloudlet. In Fig.~\ref{fig:0.5}-c, cloudlets form a communication network for exchanging necessary node features and model updates. After communicating with each other, each cloudlet constructs ST-GNN subgraph required for training on its local nodes, as shown in Fig.~\ref{fig:0.5}-d.

% talk 1st about figure 1 and then the approaches (setups)
Using this architecture, we evaluate four different training setups---centralized, traditional FL, server-free FL, and Gossip Learning---for ST-GNNs in the context of traffic prediction, with the centralized setup serving as a baseline for comparing the semi-decentralized approaches.

\subsection{GNN Partitioning in Distributed Training}
Partitioning the graph into subgraphs for local training in distributed GNN scenarios presents unique challenges due to their reliance on spatial dependencies. Unlike traditional ML models that process localized data independently, GNNs require information from neighbouring nodes to compute effective representations. Specifically, in an $\ell$-layer GNN, the representation of a node depends not only on its own features but also on the features of its $\ell$-hop neighbours.

This spatial dependency introduces a need for inter-cloudlet communication, as the required node features often extend beyond the boundaries of a single cloudlet's subgraph. Since IoT-devices typically have limited communication capabilities and cannot directly exchange data with distant cloudlets, the necessary node features are shared between the cloudlets themselves.

Each cloudlet, based on the underlying sensor network connectivity, is aware of which neighbouring cloudlets require its node features for local computations. When features are collected from the IoT-sensors, the cloudlet proactively broadcasts these features to the neighbouring cloudlets that need them for training (Fig.~\ref{fig:0.5}-d). This proactive exchange of node features ensures that each cloudlet has sufficient information to perform local computations without compromising the accuracy of the GNN model.
\section{Experimental Setup}
\subsection{Datasets}
To evaluate the performance of our setups, we utilize two real-world public traffic datasets, PeMS-BAY\footnote{\label{footnote_1}Download link: \url{https://github.com/liyaguang/DCRNN}} and METR-LA\footref{footnote_1} \cite{cite_31}, which are collected by Caltrans performance measurement system. Both datasets consist of traffic data aggregated in 5-minute intervals, with each sensor collecting 288 data points per day. The details of datasets are listed in Table~\ref{table:1}.

% TABLE I
\begin{table}[htbp]
    \centering
    \caption{Details of METR-LA and PeMS-BAY}
    \begin{tabular}{c c c c c}
        \hline
        Datasets & Nodes & Interval & TimeSteps & Attribute \\
        \hline
        METR-LA & 207 & 5 min & 34,272 & Traffic speed \\
        PeMS-BAY & 325 & 5 min & 52,116 & Traffic speed \\
        \hline
    \end{tabular}
    \label{table:1}
\end{table}

The spatial adjacency matrix is constructed from the actual road network based on distance using the formula from ChebNet \cite{cite_28}. Specifically, it encodes the spatial relationships between nodes by assigning weights to edges based on the inverse of the pairwise distances between connected nodes, effectively capturing the connectivity and proximity of locations in the traffic network. Additionally, the data were preprocessed for efficient model training and evaluation. For example, if we want to utilize the historical data spanning one hour to predict the data the next hour, we pack the sequence data in group of 12 and convert it into an instance. The dataset was then split into training, validation, and test sets in a 70:15:15 ratio. The validation set is used for early stopping and to select the best model during training. The final results are then reported on the test dataset using the best-performing model identified through validation. Additionally, to mitigate the impact of varying data magnitudes, we applied standardization to normalize the original data values, reducing the impact of the large difference in value.

\subsection{Evaluation Metrics}
In our experiments, we evaluate the performance of different setups using Weighted Mean Absolute Percentage Error (WMAPE), in addition to the widely adopted Mean Absolute Error (MAE) and Root Mean Square Error (RMSE). All metrics are computed after rescaling the predictions back to the original data range, before standardization. We use WMAPE because it accounts for the varying magnitudes of traffic speeds across regions, providing a more balanced evaluation, compared to Mean Absolute Percentage Error (MAPE). We omit the formulas for MAE and RMSE due to their wide adoption. The WMAPE formula is as follows:

\begin{equation}
    \text{WMAPE}(\bm{x}, \bm{\hat{x}}) = \frac{\sum_{i}{\mid x_i - \hat{x}_i \mid}}{\sum_{i}{\hat{x}_i}} * 100\%,
\end{equation}

\noindent where $\bm{x} = x_1,\ldots,x_n$ denotes the ground truth values of vehicle speed, and $\bm{\hat{x}} = \hat{x}_1,\ldots,\hat{x}_n$ represents their predicted values.

We compute these metrics across three forecasting horizons—--short-, mid-, and long-term---to capture fluctuations in performance over time. For centralized and FL setups, the metrics are derived from the main server's aggregated model, while for server-free FL and Gossip Learning, they are computed as a weighted average of the individual cloudlet predictions. This ensures a consistent and fair comparison across all approaches.

% Figure 2
\begin{figure}[h]
  \centering
  \includegraphics[width=1\linewidth]{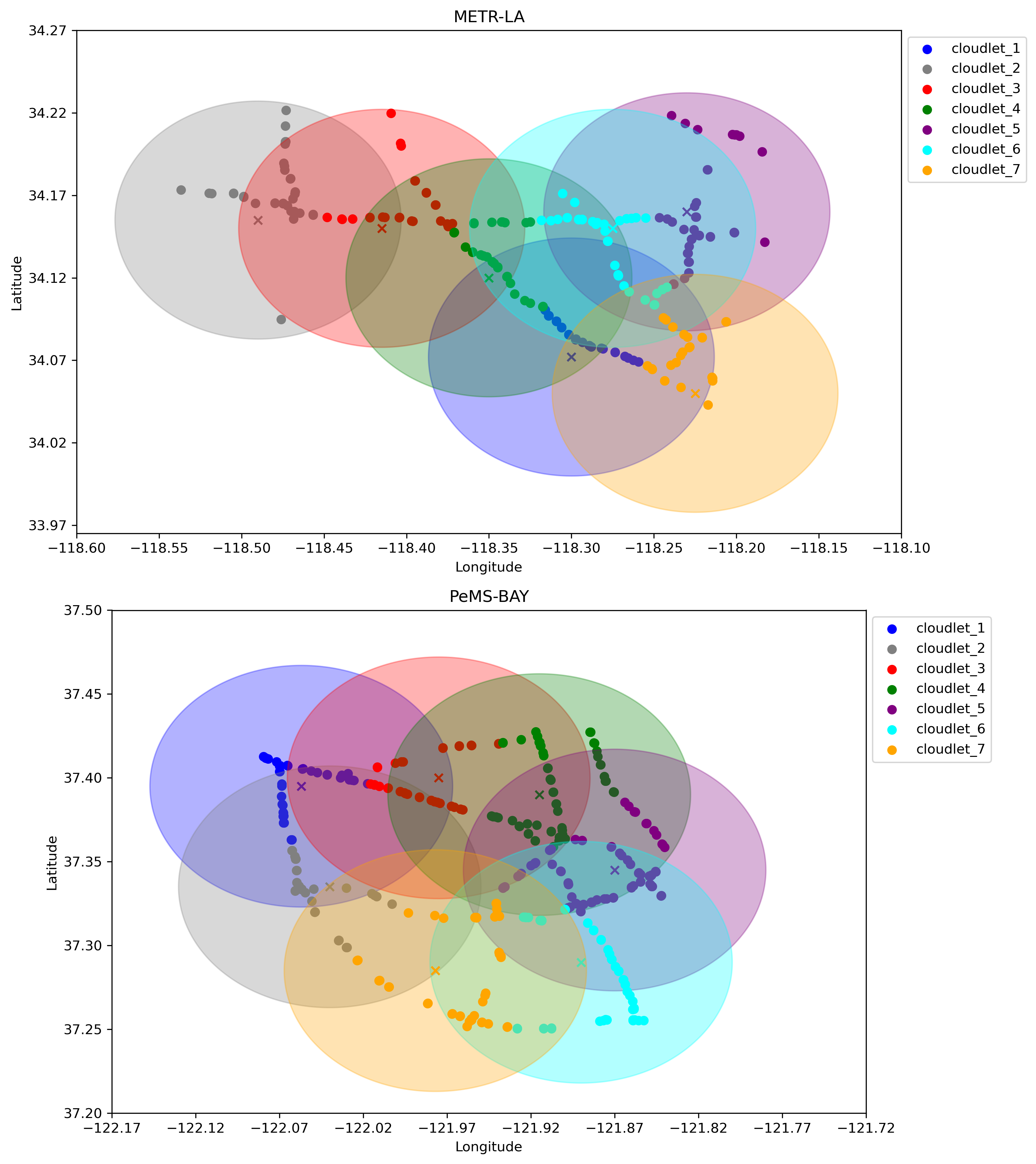}
  \caption{Sensor assignment to cloudlets based on communication range}
  \label{fig:2}
\end{figure}

\subsection{Experimental Setup}
% TABLE II
\begin{table*}[ht!]
\centering
\caption{Performance Comparison (MAE [$mile/h$]/RMSE [$mile/h$]/WMAPE [\%]) of four different setups}
\label{table:2}
\renewcommand{\arraystretch}{1.5} % Adjust row spacing for better readability
\begin{tabular}{l l ccc | ccc | ccc}
\hline\hline
\multirow{2}{*}{Dataset} & \multirow{2}{*}{Setups} & \multicolumn{3}{c|}{15 min} & \multicolumn{3}{c|}{30 min} & \multicolumn{3}{c}{60 min} \\ 
\cline{3-11}
\multicolumn{2}{l}{} & MAE & RMSE & WMAPE & MAE & RMSE & WMAPE & MAE & RMSE & WMAPE \\ 
\hline\hline
\multirow{4}{*}{METR-LA} & Centralized             & \textbf{3.78} & 9.05      & \textbf{7.45} & \textbf{5.14} & 11.61       & \textbf{10.12} & \textbf{7.35} & 14.59      & \textbf{14.47} \\ \cline{3-11}
                                      & Traditional FL      & 3.97 & 9.01       & 7.82 & 5.40 & 11.41       & 10.63 & 7.82 & 14.48      & 15.40 \\ \cline{3-11}
                                      & Server-free FL          & 3.90 & \textbf{8.98} & 7.78     & 5.35 & 11.37 & 10.67     & 7.79 & \textbf{14.41} & 15.55 \\ \cline{3-11}
                                      & Gossip Learning         & 3.88 & 9.04 & 7.74     & 5.43 & \textbf{11.35} & 10.85     & 7.56 & 14.42 & 15.10 \\ 
\hline\hline
\multirow{4}{*}{PeMS-BAY} & Centralized             & \textbf{1.48} & \textbf{3.09}      & \textbf{2.38} & \textbf{1.97} & \textbf{4.23}      & \textbf{3.17} & \textbf{2.59} & 5.41      & \textbf{4.17} \\ \cline{3-11}
                                       & Traditional FL      & 1.50 & 3.12      & 2.42 & 2.03 & 4.29      & 3.27 & 2.67 & 5.51      & 4.29 \\ \cline{3-11}
                                       & Server-free FL          & 1.50 & 3.12 & 2.42    & 2.01 & 4.25 & 3.23      & 2.65 & 5.44 & 4.28 \\ \cline{3-11}
                                       & Gossip Learning         & 1.51 & 3.12 & 2.43    & 2.03 & 4.28 & 3.26      & 2.63 & \textbf{5.39} & 4.24 \\ 
\hline\hline
\end{tabular}
\end{table*}

All experiments were run with a fixed number of 40 epochs, utilizing the High-Performance Computing (HPC) system Leonardo Booster, which features 4x NVIDIA A100 SXM6 64GB GPUs, with only 1 being used during training. The software environment used for the experiments included Python 3.10.14, PyTorch 2.3.1, and PyTorch Geometric 2.5.3.

For model architecture, we used 2 spatio-temporal blocks (ST-blocks), with GLU activation function, a learning rate of 0.0001, dropout of 0.5, weight decay of 0.00001, and a batch size of 32. MAE is used as the training loss, Adam was chosen as the optimization algorithm, and learning rate scheduler was applied with the StepLR strategy, where the step size was set to 5, and the decay factor was set to 0.7. The temporal and spatial kernel size was set to 3 for all experiments, with Chebyshev convolution as convolutional layer. All experiments use 60 minutes as the historical time window, i.e., 12 observed data points. The hyperparameters used in our experiments are based on previously established configurations that have been optimized for the given model architecture and datasets \cite{cite_13}. We adopt these settings to ensure comparability with prior work and maintain consistency in evaluating different training approaches.

Additionally, sensors were distributed across 7 cloudlets based on proximity and communication range. While algorithms such as k-means clustering could have been used to distribute sensors to cloudlets, we manually defined cloudlet placement to ensure full coverage of the geographical area in each dataset. Each cloudlet can communicate with other cloudlets and with IoT devices only if they are within an 8 km range. This limitation directly impacts the server-free FL approach by restricting the number of cloudlets that can exchange model updates. In contrast, this constraint does not affect Gossip Learning, as model updates are sent to a randomly selected cloudlet across the entire network, regardless of proximity. Fig.~\ref{fig:2} shows how sensors are assigned to cloudlets and their respective communication ranges for both the METR-LA and PeMS-BAY datasets. This partitioning reflects the semi-decentralized structure used for our distributed training approach.

% We want to see wether these dec approaches can have the same quality of perforamcne as centrlaized
% We want to look at the what are the overhead of what these dec approach bring comapred to centralize,d and wether this is a meanimgful tradeoff
% We want to look at errors at different cloudlet positioned geographicaly
Our experimental setup is designed to analyze global average prediction performance of each training approach using evaluation metrics, while also providing a comprehensive view of how errors distribute geographically across cloudlets for different prediction horizons. Additionally, we analyze the communication and computational overhead associated with each setup, with particular attention to the high cost of node feature transfer in high-density graphs.
\section{Experimental Results}
\subsection{Comparison of training setups}
Table~\ref{table:2} presents the results of all setups, datasets, and forecast horizons. Centralized training consistently outperforms semi-decentralized methods across all metrics and horizons. However, the performance gap is minimal, especially for MAE and RMSE. For instance, in the METR-LA dataset, the MAE difference between centralized and the best semi-decentralized setup is only 0.1 mph for short-term predictions—negligible compared to typical traffic flow variations, often measured in tens of mph. While WMAPE shows a slightly larger gap as the forecast horizon increases, particularly for METR-LA, the difference remains narrow across all horizons. This indicates that semi-decentralized setups offer competitive performance with no significant practical drawbacks for traffic prediction.

\subsection{Cloudlet Metric Analysis}
% Figure 3
\begin{figure*}[ht!]
  \centering
  \includegraphics[width=1\linewidth]{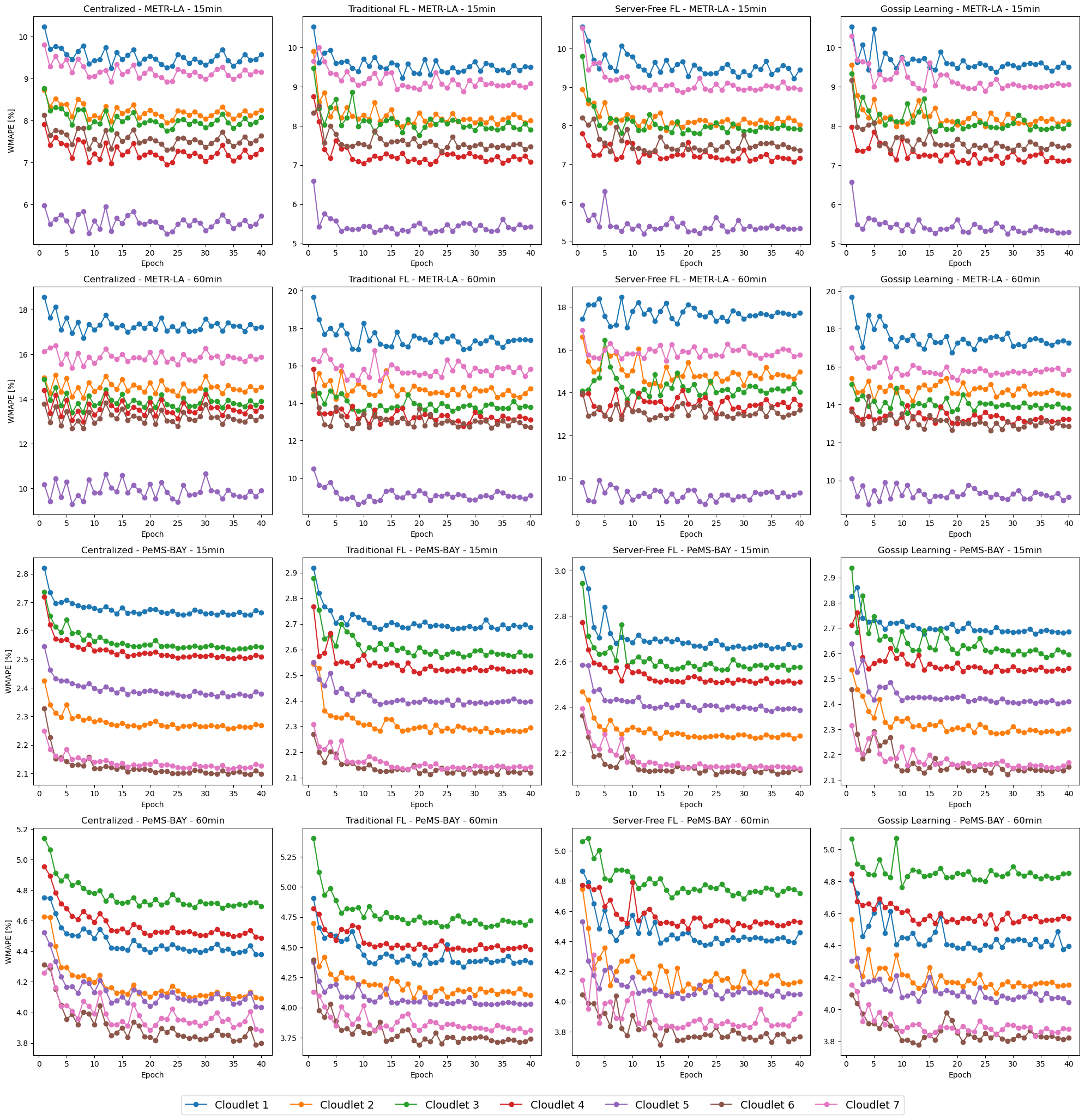}
  \caption{WMAPE for individual cloudlets}
  \label{fig:3}
\end{figure*}

Figure~\ref{fig:3} shows WMAPE variability across cloudlets for all setups and datasets, focusing on short- and long-term predictions, providing a comprehensive view of how errors distribute geographically across cloudlets for different prediction horizons. WMAPE is highlighted as it provides a clear, percentage-based measure of relative error, making it easier to compare variability across cloudlets. 

The figure reveals consistent WMAPE spread across cloudlets, regardless of training setup, dataset, or horizon, indicating that performance variability stems from the datasets' geographical characteristics and sensor partitioning, rather than the training methods. Each cloudlet's performance is influenced by the distribution of sensors assigned to it and the corresponding traffic patterns in that geographical region.

Additionally, while global weighted averages reported in Table~\ref{table:2} provide a summary of performance, they fail to capture the significant differences in performance across individual cloudlets. Previous works have primarily focused on global averages, overlooking the notable challenge of heterogeneous performance among cloudlets. This oversight can lead to an incomplete understanding of model performance in distributed traffic forecasting systems, resulting in unexpected discrepancies, suboptimal decision-making, and potential disruptions in real-world usage compared to testing.
% Figure 4
\begin{figure}[h]
  \centering
  \includegraphics[width=1\linewidth]{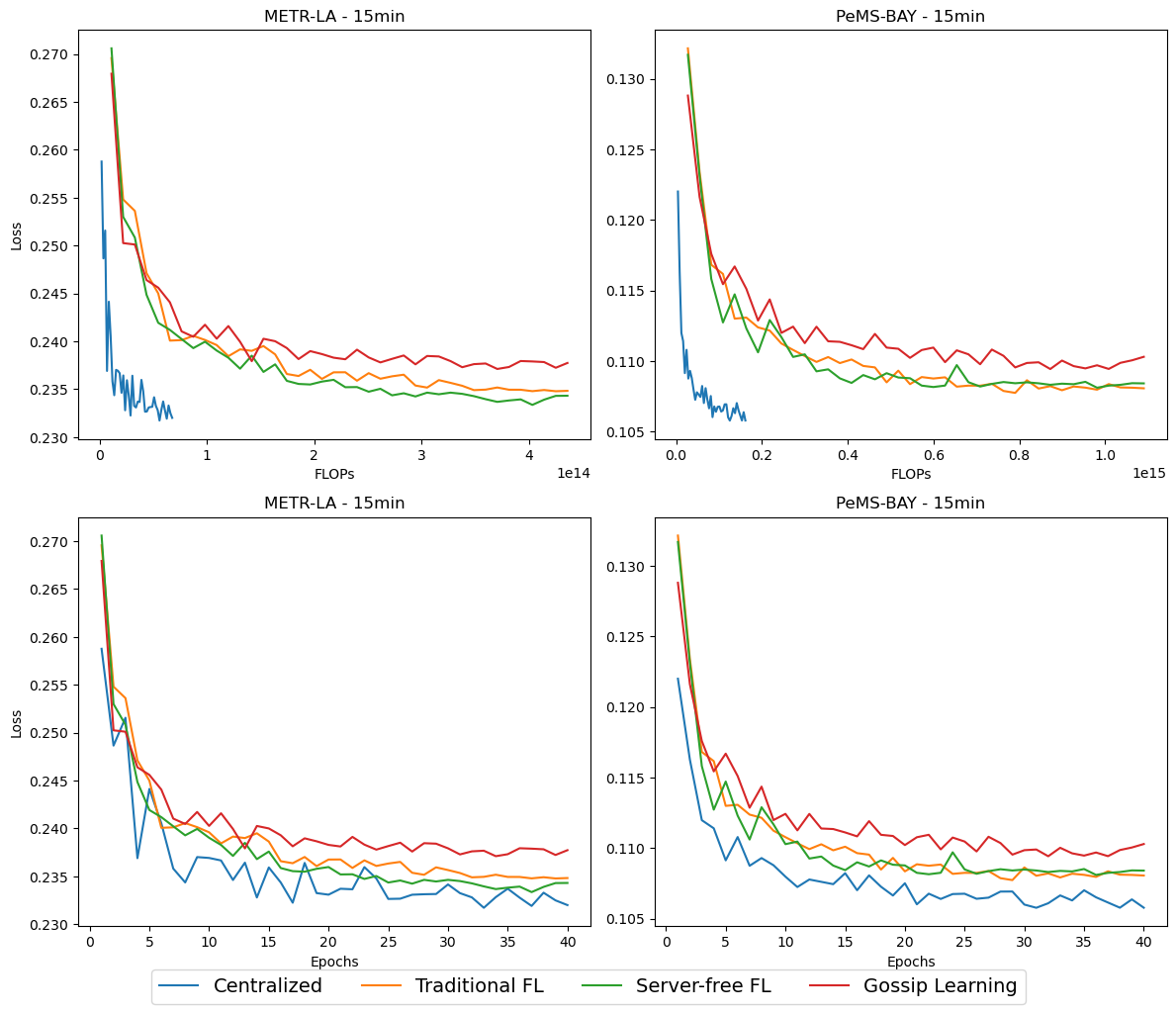}
  \caption{Validation loss over FLOPs and epochs for short-term prediction}
  \label{fig:4}
\end{figure}

% TABLE III
\begin{table*}[htbp]
    \centering
    \caption{FLOPs and average model and node feature transfer size}
    \label{table:3}
    \renewcommand{\arraystretch}{1.5} % Adjust row spacing for better readability
    \begin{tabular}{l l c | c | c | c}
    \hline\hline
    \multirow{1}{*}{Dataset} & \multirow{1}{*}{Setups} & \multicolumn{1}{c|}{Model [MB]/epoch} & \multicolumn{1}{c|}{Training FLOPs/epoch} & \multicolumn{1}{c|}{Aggregation FLOPs/epoch} & \multicolumn{1}{c}{Node feature [MB]}\\ 
    % Header row
    %\multicolumn{2}{l}{} & Model [MB] & Node feature [MB] \\ 
    \hline\hline
    % METR-LA rows
    \multirow{4}{*}{METR-LA} 
        & Centralized              & - & 1.68T & - & 4.76 \\ \cline{3-6}
        & Traditional FL & 0.91 / 7 & 10.92T / 7 & 1.12M / 7 & 25.83 / 7 \\ \cline{3-6}
        & Server-free FL           & 1.61 / 7 & 10.92T / 7 & 3.64M / 7 & 25.83 / 7 \\ \cline{3-6}
        & Gossip Learning          & 0.91 / 7 & 10.92T / 7 & 2.94M / 7 & 25.83 / 7 \\ 
    \hline\hline
    % PeMS-BAY rows
    \multirow{4}{*}{PeMS-BAY} 
        & Centralized              & - & 4.06T & - & 11.27 \\ \cline{3-6}
        & Traditional FL & 0.91 / 7 & 27.23T / 7 & 1.12M / 7 & 66.08 / 7 \\ \cline{3-6}
        & Server-free FL           & 1.82 / 7 & 27.23T / 7 & 3.92M / 7 & 66.08 / 7 \\ \cline{3-6}
        & Gossip Learning          & 0.91 / 7 & 27.23T / 7 & 2.94M / 7 & 66.08 / 7 \\ 
    \hline\hline
    \end{tabular}
\end{table*}

\subsection{Analysis of semi-decentralized overheads}
Semi-decentralized learning naturally introduces several overheads, primarily stemming from the need to communicate and aggregate models, increasing communication and computational costs compared to centralized training. Additionally, due to graph partitioning, node features must be exchanged between cloudlets, further contributing to communication and compute demands. Table~\ref{table:3} breaks down these overheads across three key aspects: model transfer, node feature transfer, and floating-point operations (FLOPs).

A significant communication cost arises from node feature transfers. Unlike centralized training, where each sensor's features are sent once to a central server, distributed setups require sending features to multiple cloudlets. This increase is due to the dense graph created by our distance-based weighted adjacency matrix and the 2-hop GNN, which requires each cloudlet to process features from both direct and indirect neighbors. Consequently, each cloudlet processes a substantial graph portion, leading to several-fold higher communication costs compared to centralized setups. However, as the network grows larger, the portion of the graph stored in each cloudlet will decrease, as cloudlets farther away will not have connected sensors. Despite the increase, the overall feature transfer per cloudlet remains manageable, typically just a few megabytes.

In contrast, model transfer per epoch is relatively small due to the compact size of ST-GCN models. While model transfers accumulate over time, contributing to total communication overhead, the per-epoch cost remains in the order of megabytes. As shown in Figure~\ref{fig:4}, distributed setups, particularly server-free FL, require more epochs to converge than centralized training, increasing total model transfer costs. This slower convergence is further influenced by the high amount of duplicated node features across cloudlets in semi-decentralized approaches. Despite this, even in server-free FL’s worst-case scenario, where all cloudlets communicate with each other, model transfer remains a minor overhead.

Computation-wise, aggregation costs are many orders of magnitude smaller training costs, so they do not represent a reason to avoid semi-decentralized training. However, distributed setups incur significantly higher training costs than centralized ones, mirroring the feature transfer pattern. Each cloudlet runs the model on its local subgraph, often with duplicated nodes and features due to overlapping receptive fields. This duplication results in additional computations, as cloudlets must compute partial embeddings for nodes they do not own. In contrast, centralized training efficiently processes the entire graph without redundancies, avoiding duplicated calculations.

Our analysis shows that the primary overheads in semi-decentralized training stem not from the distributed learning algorithms themselves, but from the highly interconnected spatial nature of traffic data and the need to partition these spatial connections across cloudlets. If these algorithms were applied to non-spatial models, the overheads would be minimal. While communication and computation costs are higher than in centralized setups, they remain within acceptable limits, making semi-decentralized training a practical solution for scalable traffic prediction in smart mobility applications.
\section{Related Work}
Training and testing GNN models on large graphs demand substantial memory and processing resources due to the interdependence of graph nodes \cite{cite_17}. This creates scalability and reliability issues for centralized GNN operations \cite{cite_18}, especially when dealing with large-scale graphs or sensitive data collected from distributed sensors in real-time traffic prediction.

\subsection{Centralized ST-GNN training}\label{AA}
However, the majority of existing approaches for training ST-GNNs are based on centralized architectures \cite{cite_12, cite_13, cite_14, cite_15}. While this approach allows for the direct management of both spatial and temporal correlations, ensuring that the full context of the data can be leveraged to improve the models accuracy and performance, it comes with significant limitations as described above.

\subsection{Traditional Federated Learning ST-GNN training}\label{AA}
Several works have applied traditional FL to train ST-GNN models for traffic prediction \cite{cite_20, cite_21, cite_22, cite_23}. One of the primary advantages of traditional FL is enhanced scalability. By distributing the training process, it leverages the computational resources of multiple devices, which allows larger datasets to be processed without overloading any single server. While this approach improves on centralized training, it still relies on a central aggregator, meaning that it shares the same issues as the centralized approach, as was already described above.

\subsection{Decentralized and semi-decentralized ST-GNN training}
To the best of our knowledge, there has been no prior research specifically focused on decentralized or even semi-decentralized training of ST-GNNs for traffic prediction.

One relevant work by L. Giaretta et. al. \cite{cite_24} explores the challenges and solutions associated with fully decentralized training. However, their solution is not tailored to ST-GNNs, but rather to Graph Convolutional Networks (GCNs), limiting its applicability to traffic prediction tasks that require both spatial and temporal capabilities.

While there is no prior research work for decentralized training for ST-GNNs, M. Nazzal et. al. \cite{cite_25} developed Heterogeneous Graph Neural Network with LSTM (HetGNN-LSTM) for taxi-demand prediction and tested it in a semi-decentralized environment, where nodes were partitioned into cloudlets, without any central aggregator. However, their solution only enables semi-decentralized inference by deploying a pre-existing model, with no support for semi-decentralized training.
\section{Future Work}
\subsection{Communication and computation overhead}
Future work should address the significant communication overhead arising from node feature transfers, which dominate the bandwidth in semi-decentralized training, and increased computation of partial embeddings during training. While communication is feasible in these setups, the cost is significant compared to centralized training. Future work should aim to shift the trade-off between scalability and reliability by reducing both network overhead and computational cost without compromising model accuracy by exploring these directions:

\begin{itemize}
    \item Changing the network connectivity to be more sparse, replacing distance-based node connectivity with road network-based connectivity, with minimal impact on the model accuracy
    \item Restricting the receptive field of GNNs by adopting layer-wise GNN training, where the receptive field is constrained to avoid unnecessary feature propagation, as proposed by L. Giaretta et. al. \cite{cite_24}.
\end{itemize}

\subsection{Cloudlet personalization}
Another promising direction is implementing individual cloudlet personalization to address performance variability across cloudlets. Our results have highlighted a notable gap in error rates, especially in certain cloudlets with consistently higher error values. Personalizing cloudlets that have lower performance compared to the rest of the cloudlets, tailoring them to local traffic conditions or unique spatial-temporal patterns, could help reduce these disparities. Such a personalized approach may involve adjusting model parameters or incorporating local fine-tuning for cloudlets with higher error rates, potentially enhancing prediction accuracy across the network while maintaining decentralized resilience.

\subsection{Cloudlet optimization}
An important limitation of our current approach is the predefined placement of cloudlets, and communication range, which has not been optimized for coverage or efficiency. The current cloudlet locations were chosen based on a simplified partitioning strategy, ensuring communication between cloudlets, and ensuring that each node is assigned to a nearby cloudlet. Future work should explore strategies for optimizing both the number and placement of cloudlets to enhance coverage and communication efficiency, while maintaining a low performance gap compared to the centralized setup, and while also leveraging the pre-existing deployment of mobile network base-stations to determine practical cloudlet placement, ensuring realistic deployment scenarios.
\section{Conclusion}
In this paper, we have explored semi-decentralized approaches for traffic prediction in the smart mobility domain, comparing centralized, traditional FL, server-free FL, and Gossip Learning setups. By distributing ST-GNN training across multiple cloudlets, we aimed to address scalability and resilience challenges associated with traditional centralized approaches.

Experimental results on the METR-LA and PeMS-BAY datasets show that semi-decentralized setups are comparable to centralized approaches in performance metrics, while offering advantages in terms of scalability and fault tolerance. We also highlight often overlooked issues in existing literature for distributed ST-GNNs, such as the variation in model performance across different geographical areas due to region-specific traffic patterns, and the significant communication overhead and computational costs that arise from the large receptive field of GNNs, leading to substantial data transfers and increased computation of partial embeddings. However, due to the planar nature of graphs, cloudlets costs remain fixed as the network grows, unlike the growing costs in a centralized approach.

\begin{comment}
    \section*{Acknowledgment}

    The preferred spelling of the word ``acknowledgment'' in America is without an ``e'' after the ``g''. Avoid the stilted expression ``one of us (R. B. G.) thanks $\ldots$''. Instead, try ``R. B. G. thanks$\ldots$''. Put sponsor acknowledgments in the unnumbered footnote on the first page.
\end{comment}

%% Define the bibliography file to be used
\bibliography{literature}

\begin{thebibliography}{10}
\expandafter\ifx\csname url\endcsname\relax
  \def\url#1{\texttt{#1}}\fi
\expandafter\ifx\csname urlprefix\endcsname\relax\def\urlprefix{URL }\fi
\expandafter\ifx\csname href\endcsname\relax
  \def\href#1#2{#2} \def\path#1{#1}\fi

\bibitem{cite_1}
C.~Bıyık, A.~Abareshi, A.~Paz, R.~A. Ruiz, R.~Battarra, C.~D. Rogers, C.~Lizarraga, Smart mobility adoption: A review of the literature, Journal of Open Innovation: Technology, Market, and Complexity 7~(2) (2021) 146.
\newblock \href {https://doi.org/https://doi.org/10.3390/joitmc7020146} {\path{doi:https://doi.org/10.3390/joitmc7020146}}.

\bibitem{cite_2}
B.~Gomes, J.~Coelho, H.~Aidos, A survey on traffic flow prediction and classification, Intelligent Systems with Applications 20 (2023) 200268.
\newblock \href {https://doi.org/https://doi.org/10.1016/j.iswa.2023.200268} {\path{doi:https://doi.org/10.1016/j.iswa.2023.200268}}.

\bibitem{cite_3}
W.~Min, L.~Wynter, Real-time road traffic prediction with spatio-temporal correlations, Transportation Research Part C: Emerging Technologies 19~(4) (2011) 606--616.

\bibitem{cite_4}
C.~Zheng, X.~Fan, C.~Wang, J.~Qi, Gman: A graph multi-attention network for traffic prediction, in: Proceedings of the AAAI conference on artificial intelligence, Vol.~34, 2020, pp. 1234--1241.

\bibitem{cite_5}
R.~Vinayakumar, K.~Soman, P.~Poornachandran, Applying deep learning approaches for network traffic prediction, in: 2017 International Conference on Advances in Computing, Communications and Informatics (ICACCI), IEEE, 2017, pp. 2353--2358.

\bibitem{cite_6}
A.~Abadi, T.~Rajabioun, P.~A. Ioannou, Traffic flow prediction for road transportation networks with limited traffic data, IEEE transactions on intelligent transportation systems 16~(2) (2014) 653--662.

\bibitem{cite_34}
B.~M. Williams, L.~A. Hoel, Modeling and forecasting vehicular traffic flow as a seasonal arima process: Theoretical basis and empirical results, Journal of transportation engineering 129~(6) (2003) 664--672.

\bibitem{cite_35}
C.~Bent{\'e}jac, A.~Cs{\"o}rg{\H{o}}, G.~Mart{\'\i}nez-Mu{\~n}oz, A comparative analysis of gradient boosting algorithms, Artificial Intelligence Review 54 (2021) 1937--1967.

\bibitem{cite_7}
N.~Ramakrishnan, T.~Soni, Network traffic prediction using recurrent neural networks, in: 2018 17th IEEE International Conference on Machine Learning and Applications (ICMLA), IEEE, 2018, pp. 187--193.

\bibitem{cite_8}
D.~Andreoletti, S.~Troia, F.~Musumeci, S.~Giordano, G.~Maier, M.~Tornatore, Network traffic prediction based on diffusion convolutional recurrent neural networks, in: IEEE INFOCOM 2019-IEEE Conference on Computer Communications Workshops (INFOCOM WKSHPS), IEEE, 2019, pp. 246--251.

\bibitem{cite_29}
Z.~Wu, S.~Pan, F.~Chen, G.~Long, C.~Zhang, P.~S. Yu, \href{http://dx.doi.org/10.1109/TNNLS.2020.2978386}{A comprehensive survey on graph neural networks}, IEEE Transactions on Neural Networks and Learning Systems 32~(1) (2021) 4–24.
\newblock \href {https://doi.org/10.1109/tnnls.2020.2978386} {\path{doi:10.1109/tnnls.2020.2978386}}.
\newline\urlprefix\url{http://dx.doi.org/10.1109/TNNLS.2020.2978386}

\bibitem{cite_16}
Z.~A. Sahili, M.~Awad, Spatio-temporal graph neural networks: A survey (2023).
\newblock \href {http://arxiv.org/abs/2301.10569} {\path{arXiv:2301.10569}}.

\bibitem{cite_31}
Y.~Li, R.~Yu, C.~Shahabi, Y.~Liu, \href{https://arxiv.org/abs/1707.01926}{Diffusion convolutional recurrent neural network: Data-driven traffic forecasting} (2018).
\newblock \href {http://arxiv.org/abs/1707.01926} {\path{arXiv:1707.01926}}.
\newline\urlprefix\url{https://arxiv.org/abs/1707.01926}

\bibitem{cite_36}
S.~Nuli, N.~Vikranth, K.~A. Gupta, Real-time traffic prediction using neural networks, in: IOP Conference Series: Earth and Environmental Science, Vol. 1086, IOP Publishing, 2022, p. 012029.

\bibitem{cite_12}
L.~Ruiz, F.~Gama, A.~Ribeiro, Gated graph recurrent neural networks, IEEE Transactions on Signal Processing 68 (2020) 6303–6318.
\newblock \href {https://doi.org/10.1109/tsp.2020.3033962} {\path{doi:10.1109/tsp.2020.3033962}}.

\bibitem{cite_13}
B.~Yu, H.~Yin, Z.~Zhu, Spatio-temporal graph convolutional networks: A deep learning framework for traffic forecasting, in: Proceedings of the Twenty-Seventh International Joint Conference on Artificial Intelligence, IJCAI-2018, International Joint Conferences on Artificial Intelligence Organization, 2018.
\newblock \href {https://doi.org/10.24963/ijcai.2018/505} {\path{doi:10.24963/ijcai.2018/505}}.

\bibitem{cite_14}
T.~Wu, F.~Chen, Y.~Wan, Graph attention lstm network: A new model for traffic flow forecasting, in: 2018 5th International Conference on Information Science and Control Engineering (ICISCE), 2018, pp. 241--245.
\newblock \href {https://doi.org/10.1109/ICISCE.2018.00058} {\path{doi:10.1109/ICISCE.2018.00058}}.

\bibitem{cite_30}
P.~Kairouz, H.~B. McMahan, B.~Avent, A.~Bellet, M.~Bennis, A.~N. Bhagoji, K.~Bonawitz, Z.~Charles, G.~Cormode, R.~Cummings, et~al., Advances and open problems in federated learning, Foundations and trends{\textregistered} in machine learning 14~(1--2) (2021) 1--210.

\bibitem{cite_26}
C.~He, C.~Tan, H.~Tang, S.~Qiu, J.~Liu, Central server free federated learning over single-sided trust social networks, arXiv preprint arXiv:1910.04956 (2019).

\bibitem{cite_27}
R.~Orm{\'a}ndi, I.~Heged{\H{u}}s, M.~Jelasity, Gossip learning with linear models on fully distributed data, Concurrency and Computation: Practice and Experience 25~(4) (2013) 556--571.

\bibitem{cite_25}
M.~Nazzal, A.~Khreishah, J.~Lee, S.~Angizi, A.~Al-Fuqaha, M.~Guizani, Semi-decentralized inference in heterogeneous graph neural networks for traffic demand forecasting: An edge-computing approach, IEEE Transactions on Vehicular Technology (2024).

\bibitem{cite_28}
M.~Defferrard, X.~Bresson, P.~Vandergheynst, Convolutional neural networks on graphs with fast localized spectral filtering, Advances in neural information processing systems 29 (2016).

\bibitem{cite_17}
D.~Zheng, C.~Ma, M.~Wang, J.~Zhou, Q.~Su, X.~Song, Q.~Gan, Z.~Zhang, G.~Karypis, Distdgl: Distributed graph neural network training for billion-scale graphs (2021).
\newblock \href {http://arxiv.org/abs/2010.05337} {\path{arXiv:2010.05337}}.

\bibitem{cite_18}
L.~Zeng, C.~Yang, P.~Huang, Z.~Zhou, S.~Yu, X.~Chen, Gnn at the edge: Cost-efficient graph neural network processing over distributed edge servers (2022).
\newblock \href {http://arxiv.org/abs/2210.17281} {\path{arXiv:2210.17281}}.

\bibitem{cite_15}
Z.~Diao, X.~Wang, D.~Zhang, Y.~Liu, K.~Xie, S.~He, Dynamic spatial-temporal graph convolutional neural networks for traffic forecasting, Proceedings of the AAAI Conference on Artificial Intelligence 33~(01) (2019) 890--897.
\newblock \href {https://doi.org/10.1609/aaai.v33i01.3301890} {\path{doi:10.1609/aaai.v33i01.3301890}}.

\bibitem{cite_20}
L.~Liu, Y.~Tian, C.~Chakraborty, J.~Feng, Q.~Pei, L.~Zhen, K.~Yu, Multilevel federated learning-based intelligent traffic flow forecasting for transportation network management, IEEE Transactions on Network and Service Management 20~(2) (2023) 1446--1458.
\newblock \href {https://doi.org/10.1109/TNSM.2023.3280515} {\path{doi:10.1109/TNSM.2023.3280515}}.

\bibitem{cite_21}
T.~Qi, L.~Chen, G.~Li, Y.~Li, C.~Wang, Fedagcn: A traffic flow prediction framework based on federated learning and asynchronous graph convolutional network, Applied Soft Computing 138 (2023) 110175.
\newblock \href {https://doi.org/https://doi.org/10.1016/j.asoc.2023.110175} {\path{doi:https://doi.org/10.1016/j.asoc.2023.110175}}.

\bibitem{cite_22}
X.~Yuan, J.~Chen, J.~Yang, N.~Zhang, T.~Yang, T.~Han, A.~Taherkordi, Fedstn: Graph representation driven federated learning for edge computing enabled urban traffic flow prediction, IEEE Transactions on Intelligent Transportation Systems 24~(8) (2023) 8738--8748.
\newblock \href {https://doi.org/10.1109/TITS.2022.3157056} {\path{doi:10.1109/TITS.2022.3157056}}.

\bibitem{cite_23}
M.~Xia, D.~Jin, J.~Chen, Short-term traffic flow prediction based on graph convolutional networks and federated learning, IEEE Transactions on Intelligent Transportation Systems 24~(1) (2023) 1191--1203.
\newblock \href {https://doi.org/10.1109/TITS.2022.3179391} {\path{doi:10.1109/TITS.2022.3179391}}.

\bibitem{cite_24}
L.~Giaretta, S.~Girdzijauskas, Fully-decentralized training of gnns using layer-wise self-supervision (2023).

\end{thebibliography}

\end{document}